\documentclass{article}

\PassOptionsToPackage{numbers, compress}{natbib}

\usepackage[preprint]{neurips_2026}

\usepackage[utf8]{inputenc} 
\usepackage[T1]{fontenc}    
\usepackage{hyperref}       
\usepackage{url}            
\usepackage{booktabs}       
\usepackage{amsfonts}       
\usepackage{nicefrac}       
\usepackage{microtype}      
\usepackage{xcolor}         
\usepackage{graphicx}
\usepackage[table]{xcolor}
\usepackage{multirow}
\usepackage{caption}
\usepackage[export]{adjustbox}
\usepackage{amsmath}
\usepackage{float}
\usepackage{tcolorbox}

\title{Sparse Autoencoders for Interpretable Out-of-Distribution Detection}

\author{%
  Ayush Karmacharya$^{*1}$, Luke Luschwitz$^{*1}$, Lucia Romero$^{1}$, Yanan Niu$^{2}$, Joseph Campbell$^{1}$ \\
  $^{1}$ Purdue University \quad $^{2}$ EPFL\\
  \texttt{\{akarmach, lluschwi, romer190, joecamp\}@purdue.edu, yanan.niu@epfl.ch} \\
  $^{*}$ Equal contribution
}

\begin{document}

\newcolumntype{s}{>{\columncolor{gray!15}} c}

\maketitle

\begin{abstract}
Reliable detection of out-of-distribution (OOD) samples is crucial for the safe deployment of machine learning models. Neural networks often produce overconfident predictions for inputs that deviate from their training data, leading to significant degradation in performance. While many OOD detection methods focus on the final output layer, they neglect the rich hierarchical information present in intermediate network layers. This paper introduces a novel approach that leverages sparse autoencoders (SAEs) to learn interpretable features from these intermediate activations. We find that in-distribution (ID) and OOD data activate distinct sets of these sparse features. We propose a new OOD score derived from the cosine similarity between the sparse feature activations of a test sample and the mean activations of ID classes. Our post-hoc detection method not only achieves state-of-the-art performance on standard OOD detection benchmarks, but yields interpretable insights into how distribution shift affects learned representations.
\end{abstract}

\section{Introduction}

Machine learning algorithms are built on the assumption that data encountered at test-time comes from the same distribution as seen during training.
When this assumption is violated, test data is said to be out-of-distribution (OOD) and prediction accuracy rapidly degrades~\cite{moreno2012unifying}, particularly for deep neural networks (DNNs) which are prone to overconfident predictions~\cite{guo2017calibrationmodernneuralnetworks}.
However, because DNNs are effectively ``black boxes'', it is difficult to assess when their predictions can be trusted~\cite{lipton2018mythos}.

\begin{figure}[t]
\centering
\includegraphics[width=1\columnwidth]{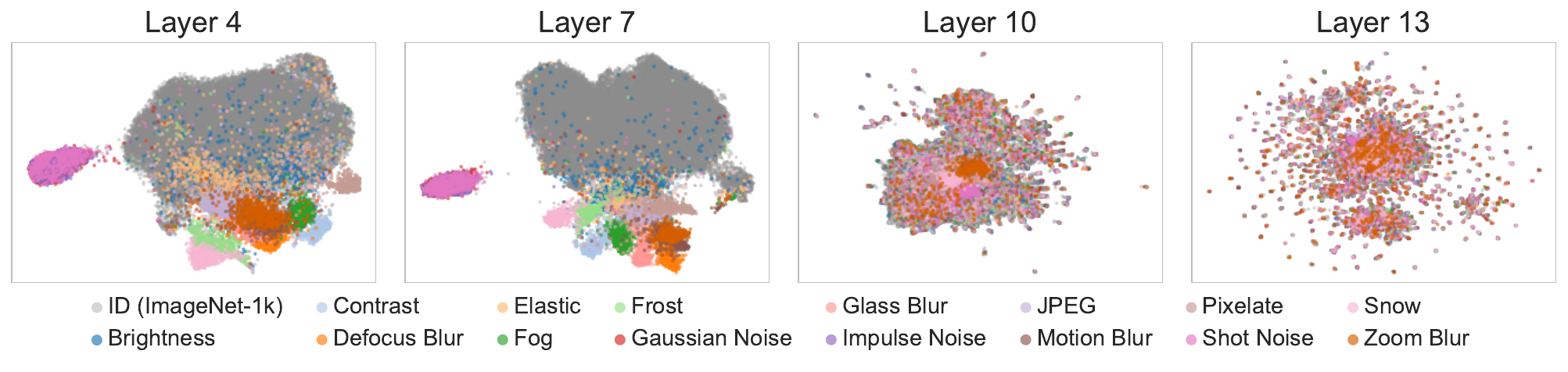} 
\caption{
UMAP~\cite{mcinnes2018umap} projections of SAE embeddings from successive layers of a vision transformer (ViT)~\cite{dosovitskiy2021imageworth16x16words} model trained on ImageNet-1k.
Points represent ImageNet-1k samples as ID and ImageNet-C samples as covariate-shifted OOD inputs, with ImageNet-C samples colored by corruption type.
In the final layer, corrupted samples largely overlap with the ID manifold, making the shift difficult to distinguish qualitatively.
In contrast, earlier layers, especially layers 4 and 7, show clearer separation between ID and OOD samples and reveal clusters corresponding to different corruption types.
This illustrates that discriminative structure for covariate-shift detection can appear in intermediate representations, motivating layer-wise OOD detection methods such as SAID.
}
\label{fig:intro}
\end{figure}

In this work, we address the problem of OOD detection~\cite{yang2024generalized}: classifying test-time inputs as either in-distribution (ID) or OOD.
Effective OOD detection is critical for real-world deployment, as models are frequently exposed to inputs that deviate from their training distribution for which they should avoid making predictions.
Yet, current OOD detection methods exhibit two significant limitations.
First, these methods typically rely almost exclusively on the final-layer representation of the network (e.g., logits), disregarding the rich, hierarchical representations~\cite{chen2021understandinghierarchicallearningbenefits} present throughout the intermediate layers~\cite{hendrycks2018baselinedetectingmisclassifiedoutofdistribution, liu2021energybasedoutofdistributiondetection, sun2022outofdistributiondetectiondeepnearest, lee2018simpleunifiedframeworkdetecting, djurisic2023extremelysimpleactivationshaping}.
As we illustrate in Fig.~\ref{fig:intro}, this can overlook discriminative structure that is visible earlier in the network but becomes less separable in later representations.
Second, even when OOD detectors perform well, their scores typically provide limited insight into the internal representations that distinguish ID from OOD inputs.
This makes it difficult to diagnose where distributional differences emerge within a network, and whether they are captured by low-level, mid-level, or high-level features.
Such a layer-wise view can inform which representations should be monitored for OOD evidence and whether a shift reflects low-level visual corruption, higher-level semantic mismatch, or both.

To address these challenges, we propose \textbf{S}parse \textbf{A}utoencoders for \textbf{I}nterpretable \textbf{D}etection (SAID), a post-hoc OOD detection method that leverages sparse autoencoders (SAEs) trained on intermediate network activations.
SAEs have recently been used to extract sparse, concept-like features from otherwise opaque neural representations~\cite{cunningham2023sparse}. 
We adapt this idea to vision-based OOD detection by training SAEs on selected layers of a frozen image classifier, including both convolutional and Transformer-based vision models.
At test time, SAID compares the SAE encoding of an input to the mean SAE encoding of the predicted ID class, producing a similarity-based OOD score.
We show that these sparse intermediate representations improve detection performance across both semantic and covariate distribution shifts.

Beyond detection performance, the sparse structure of SAE representations provides a diagnostic lens for analyzing how ID and OOD inputs activate the model’s learned concept space.
We automatically annotate SAE latents with human-readable concept labels using a vision-language model, then evaluate whether the concepts activated by a test input are semantically consistent with the image.
Our results show that ID samples tend to exhibit higher concept agreement than both semantic- and covariate-shift OOD samples, and that this agreement gap varies across layers.


Our contributions are threefold. First, we introduce SAID, a post-hoc OOD detection method that uses SAE encodings of intermediate-layer representations to improve detection performance.
Second, we show that intermediate layers contain discriminative information that is often lost or compressed in final-layer representations, improving performance across semantic and covariate shift benchmarks.
Third, we analyze SAE-derived concept labels as a diagnostic tool for understanding how distribution shift affects learned representations, showing that ID inputs activate concepts that are more semantically consistent with the image than OOD inputs.

\section{Related Work}


\textbf{Out-of-Distribution Detection.}
One prominent category of OOD detection methods leverages the uncertainty of model predictions or the logits (outputs of the final layer). For instance, softmax confidence scores have been used as a baseline for OOD detection \cite{hendrycks2018baselinedetectingmisclassifiedoutofdistribution}. Building upon this, energy-based models have been proposed, where the energy score derived from the model's output logits is used to distinguish between in-distribution and OOD samples \cite{liu2021energybasedoutofdistributiondetection}. While these methods offer a starting point, they primarily rely on information from the final layer, potentially overlooking valuable insights embedded within the intermediate representations of the neural network \cite{guglielmo2025leveragingintermediaterepresentationsforbetterout-of-distributiondetection}.

To overcome the limitations of solely relying on final layer outputs, some research has explored using features from earlier layers. The Mahalanobis distance applied to intermediate feature representations has shown promise in improving near-OOD detection \cite{lee2018simpleunifiedframeworkdetecting, ren2021simplefixmahalanobisdistance}. Similarly, methods utilizing deep nearest neighbors in feature space have been developed for OOD detection \cite{sun2022outofdistributiondetectiondeepnearest}. More recent advancements include activation shaping techniques, which aim to directly influence the activations of neural networks to improve OOD discrimination \cite{djurisic2023extremelysimpleactivationshaping}. Despite these advancements, a challenge remains in effectively utilizing the vast amount of information contained in all intermediate activations, as it can be redundant, noisy, or difficult to interpret.

\textbf{Interpretable Machine Learning.}
The increasing interest in interpretable machine learning has led to methods that explain model
predictions using human-understandable concepts
\cite{ribeiro2016whyitrustyou, choi2023conceptbasedexplanationsoutofdistributiondetectors}.
This is especially relevant for OOD detection, where concept-level representations can help analyze
how ID and OOD inputs differ. Recent work has also begun to explore the role of feature sparsity in
OOD detection \cite{Chen_Li_Chen_Maul_Yin_2024}.

Sparse Autoencoders (SAEs) provide a natural mechanism for learning sparse, interpretable features
from neural network activations
\cite{makhzani2014ksparseautoencoders, elhage2022toymodelssuperposition,
bricken2023monosemanticity, cunningham2023sparseautoencodershighlyinterpretable}.
By decomposing dense activations into localized, concept-like features
\cite{gao2024scalingevaluatingsparseautoencoders, lieberum2024gemmascopeopensparse},
SAEs offer a way to analyze otherwise opaque representations and have connections to anomaly
detection
\cite{sarvari2019unsupervisedboostingbasedautoencoderensembles,
sun2018learningsparserepresentationwithvariationalauto-encoderforanomalydetection}.
Universal SAEs further suggest the possibility of aligning sparse concepts across models
\cite{thasarathan2025universalsparseautoencodersinterpretable}.

Building on this, we use SAEs to extract sparse concept representations from intermediate layers for
OOD detection. Rather than relying only on final-layer features, our approach leverages the distinct
activation patterns induced by ID and OOD inputs throughout the network, providing both strong OOD
detection and a concept-level view of distribution shift.

\section{Preliminaries}

\subsection{Out-of-Distribution Detection}

Let a machine learning model be trained on an ID dataset, $\mathcal{D}_{in} = \{ (\mathbf{x}_i, y_i) \}_{i=1}^N$, where $\mathbf{x}_i \in \mathbf{R}^d$ are input data points and $y_i$ are their corresponding labels.
We formulate the task of OOD detection as a binary classification problem: given a test sample $\mathbf{x}$, the goal is to determine whether it is drawn from the same distribution as the training set (ID) or from a different distribution (OOD).
In this work, we focus on two causes of distributional shift that can result in data being OOD~\cite{yang2024generalized}. The first is \textit{semantic shift}, where the label distribution $p(y|x)$ of the test distribution differs from training. The second is the problem of \textit{covariate shift}, where the distribution over the inputs, $p(x)$, differs.

\subsection{Sparse Autoencoders}

An autoencoder is a neural network designed to learn a compressed representation (encoding) of input data. It is comprised of an encoder, $E(\cdot)$, which maps an input $\mathbf{x}$ to a latent representation (or code) $\mathbf{z}$, and a decoder, $D(\cdot)$, which reconstructs the input from this latent representation, aiming for $D(\mathbf{z}) \approx \mathbf{x}$. The primary objective during training is to minimize the mean squared error between the original input and its reconstructed version plus an $\mathcal{L}_1$ term to induce sparsity.

A sparse autoencoder introduces a sparsity constraint on the latent representation $\mathbf{z}$. This encourages most of the units in the hidden (latent) layer to be inactive (i.e., have outputs very close to zero) for any given input, ensuring that only a small, specific subset of features are ``active."

In this work, we specifically utilize a top-$k$ sparse autoencoder. For a given input $\mathbf{a}$ (representing an intermediate activation from a pre-trained neural network), the encoder $E(\cdot)$ first computes an intermediate latent vector. This vector is then subjected to a top-$k$ activation function, which explicitly forces all but the $k$ most active latent dimensions to zero. Top-$k$ SAEs do not have the $\mathcal{L}_1$ sparsity term in the objective function.

This direct method of enforcing sparsity encourages the learned features to be monosemantic, meaning each feature in the sparse representation ideally corresponds to a single, distinct concept. This property is  suited for interpretability of the learned features, providing insight into what specific concepts the network is activating for a given input.

\begin{figure*}[t]
\centering
\includegraphics[width=0.99\textwidth]{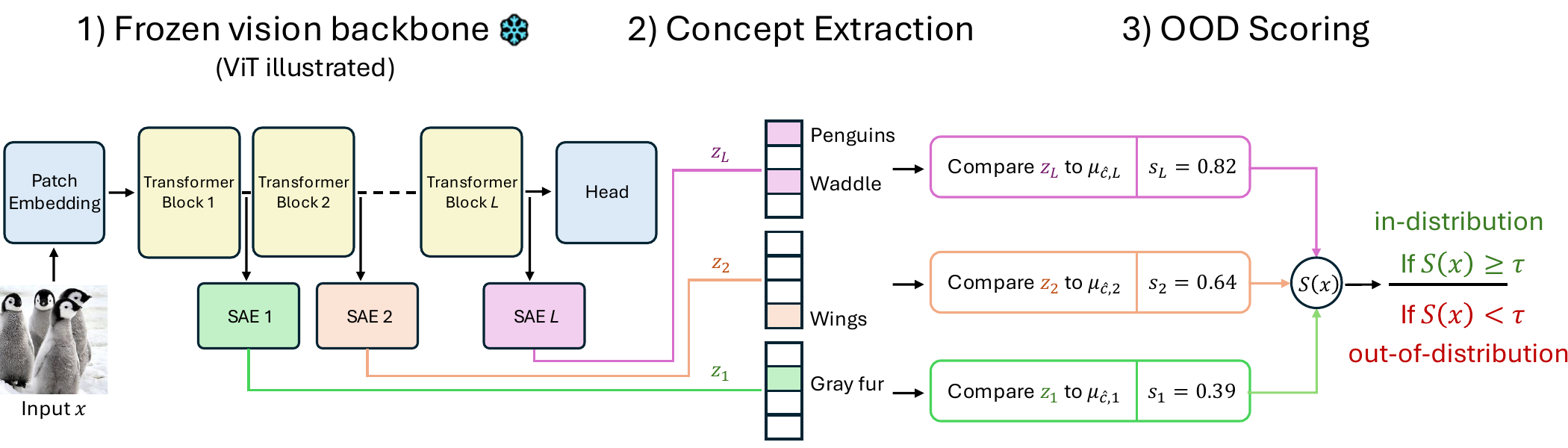} 
\caption{Sparse autoencoders are trained on the activations of intermediate layers of a frozen model over an in-distribution dataset. At test-time, a given sample $\mathbf{x}$ is considered OOD if the average cosine similarities between its latent encoding $z_l$ and the corresponding mean latent encoding for the predicted class $\mu_{\hat{c}}$ over all layers is below a threshold $\tau$.}
\label{fig:overview}
\end{figure*}

\section{Sparse Autoencoders for Interpretable Detection}

We introduce a \textit{post-hoc} method for OOD detection which leverages the latent encodings from SAEs.
Our approach, \textbf{S}parse \textbf{A}utoencoders for \textbf{I}nterpretable \textbf{D}etection (SAID), trains SAEs on the intermediate feature activations of a pre-trained vision-based classification model, as shown in Fig.~\ref{fig:overview}.

\subsection{Training Algorithm}

Given a pre-defined set of layers $L$ we train one SAE for each layer $l\in L$ of a pre-trained neural network $f(\cdot)$.
For each layer, the corresponding SAE receives as input the intermediate activations $\mathbf{a}=f_l(\mathbf{x})$, where $f_l$ is the $l$-th layer of $f$.
All SAEs use identical hyperparameters, excluding their sparsity coefficient $k$.
To construct the SAE training set, we pass the entire original dataset through the pre-trained network $f$ and store the resulting activations from each intermediate layer $l \in L$.
Each SAE is trained to minimize a mean squared error (MSE) reconstruction term. Formally, for a batch of intermediate activations $\mathbf{a}$, the MSE reconstruction loss is defined as follows:

\[ \mathcal{L}_{\mathrm{MSE}} = \frac{1}{N} \sum_{i=1}^{N} \|\mathbf{a}^{(i)} - D(E(\mathbf{a}^{(i)}))\|_2^2 .\]
where $E$ and $D$ denote the SAE encoder and decoder respectively, and $N$ the number of samples in the batch. After training, we freeze the SAE and only use its encoder $E_l$ for OOD scoring.

\subsection{Similarity-Based Scoring}

After training each SAE for a given layer $l \in L$, we encode all training samples from each ID class into the corresponding sparse latent space.
We then compute and store the mean latent encoding $\mu_{{c},l}$ for every ID class $c$.
Thus, the total number of stored class means equals the number of ID classes multiplied by the number of SAEs (one per layer).

For a test input $\mathbf{x}$, let $\hat{c} = \arg\max_c f_c(\mathbf{x})$ denote the class predicted by the pre-trained classifier, where $f_c(\mathbf{x})$ denotes the predicted probability assigned to class $c$. During inference, each SAE produces a latent encoding $\mathbf{z}_l =E_l(f_l(\mathbf{x}))$.
We compute a layer-specific OOD score $s_l(\mathbf{x})$ as the cosine similarity between this latent encoding and the mean encoding of the predicted class $\mu_{\hat{c},l}$:
\[ s_l(\mathbf{x})
 = \frac{\mathbf{z}_l \cdot \mu_{\hat{c},l}}{\|\mathbf{z}_l\| \|\mu_{\hat{c},l}\|}.\]

These layer-wise scores are then aggregated into a single unified OOD confidence score $S(\mathbf{x})$ by taking the average over all layer scores for a given sample. Thus, the overall OOD score $S$ is computed as follows,

\[
S(\mathbf{x}) = \frac{1}{|L|}\sum_{l \in L}  s_l(\mathbf{x}).
\]

At test-time, we apply a threshold $\tau$ to classify samples: inputs for which $S(\mathbf{x}) \geq \tau$ are classified as ID, while inputs with $S(\mathbf{x}) < \tau$ are classified as OOD.
Intuitively, higher scores indicate greater similarity between the sample and the known ID distribution.

\subsection{VLM-Based Concept Annotation}

A key benefit of using SAEs is their ability to learn interpretable features.
To leverage this interpretability for concept-based explanations of OOD predictions, we introduce an automated labeling process guided by a vision-language model (VLM).
Following strategies from natural language domains~\cite{bills2023language}, we use a pre-trained VLM (GPT-5.5~\cite{openai2026gpt55}) to assign human-interpretable labels to latent features.

For each feature in the sparse latent space of each SAE, we identify the top 25 images from the ID training set which activate that latent most strongly, sorted by activation magnitude.
These images represent the visual stimuli most closely associated with that latent concept.
We then pass this set of images to a pretrained VLM along with a structured prompt (provided in the Appendix) that instructs the VLM to describe the shared visual patterns or semantic theme of these images, yielding a concise, human-readable concept label.

Fig.~\ref{fig:concepts} shows examples of VLM-generated annotations, alongside their representative sets of images.
Although each SAE theoretically possesses as many concepts as there are latent features, in practice, the presence of dead latents reduces the effective number of concepts.
These labels provide an explanation of why specific inputs are considered out-of-distribution, highlighting precisely which concept-based features are or are not activated.

\section{Experiments}


We evaluate our proposed approach, SAID, with the goal of answering three questions: 1) Is SAID effective at detecting both semantic and covariate shifts in data across multiple architectures? 2) Does using the intermediate layers of a network provide added benefit over just using the final layer of a network? 3) Does SAID's concept-based explanations reflect meaningful semantic differences between ID and OOD inputs? 

\textbf{Datasets.} We evaluate our performance using ImageNet-1K~\cite{imagenet15russakovsky} as the ID dataset and select OOD datasets containing both semantic and covariate distribution shifts.
Semantic shifted datasets include: SSB-Hard~\cite{vaze2022openset}, NINCO~\cite{bitterwolf2023outfixingimagenetoutofdistribution}, iNaturalist~\cite{van2018inaturalist}, Textures~\cite{cimpoi14describing}, OpenImage-O~\cite{wang2022vimoutofdistributionvirtuallogitmatching}, MNIST~\cite{deng2012mnist}, FashionMNIST~\cite{xiao2017fashion}, and COVID~\cite{winther2020covid19_image_repository}.
To study covariate shifts, we use the following ImageNet variants: ImageNet-C~\cite{hendrycks2019benchmarkingneuralnetworkrobustness}, ImageNet-ES~\cite{baek2024unexplored}, and ImageNet-R~\cite{hendrycks2021facesrobustnesscriticalanalysis}.

We test our method on two different, publicly available architectures pretrained on ImageNet-1K. The first is a transformer-based architecture, ViT-B/16~\cite{dosovitskiy2021imageworth16x16words}, and the second is the CNN-based ResNet-50~\cite{he2015deepresiduallearningimage}.
Unlike methods which perform training-time regularization, our method does not modify the parameters of the backbone model.
Thus, we only compare our method to other post-hoc detection methods.

\textbf{Baselines.} We compare our method to seven baseline methods: 
VIM~\cite{wang2022vimoutofdistributionvirtuallogitmatching}, RMDS~\cite{ren2021simplefixmahalanobisdistance}, MDS~\cite{lee2018simpleunifiedframeworkdetecting}, KNN~\cite{sun2022outofdistributiondetectiondeepnearest}, EBO~\cite{liu2021energybasedoutofdistributiondetection}, MLS~\cite{pmlr-v162-hendrycks22a}, and SHE~\cite{zhang2023outofdistribution}. These implementations were taken from OpenOOD~\cite{zhang2024openoodv15enhancedbenchmark}, a library for benchmarking OOD performance.

\textbf{Evaluation Metrics.} We evaluate our method using two metrics. AUROC measures the method's overall ability to distinguish ID from OOD samples across all thresholds, with 100 indicating perfect detection and 50 equivalent to random guessing. FPR95 measures the percentage of OOD samples incorrectly identified as ID when the true positive rate for ID samples is set to 95\%; lower values indicate better performance.

All details regarding hyperparameters, training and evaluation details, and computational resources can be found in the Appendix.

\begin{table*}[h]
\caption{\textbf{Semantic shift} detection results for SAID and baseline methods across two models: ViT and ResNet-50. We use ImageNet-1k as the ID dataset and report AUROC $\uparrow$ / FPR95 $\downarrow$ values for nine OOD datasets exhibiting semantic shift. Bold indicates the best method.}
\label{tab:semantic_shift_results}
\centering
\scriptsize
\setlength{\tabcolsep}{3pt}
\renewcommand{\arraystretch}{1.05}
\begin{tabular}{|c|c|cccccccc|c|}
    \hline
     & Method & SSB-Hard & NINCO & iNaturalist & Textures & OpenImage-O & MNIST & FashionMNIST & COVID & Average \\
    \hline

    \multirow{8}{*}{\rotatebox[origin=c]{90}{ViT}}
    & VIM & 70.7 / 89.6 & 85.7 / 60.3 & \textbf{97.2} / \textbf{11.6} & 90.2 / 42.4 & 92.4 / 30.7 & 92.7 / 15.8 & 89.5 / 27.1 & 96.1 / 10.4 & 89.3 / 36.0 \\
    & RMDS & \textbf{72.6} / 85.0 & 87.2 / \textbf{46.9} & 96.1 / 19.8 & 89.3 / 37.6 & 92.2 / 29.8 & 91.4 / 18.5 & 91.9 / 20.9 & 91.6 / 28.0 & 89.0 / 35.8 \\
    & MDS & 71.3 / 84.0 & 86.4 / 48.9 & 96.0 / 20.7 & 89.3 / 39.3 & 92.3 / 30.6 & 93.8 / 13.6 & 93.4 / 17.6 & 92.6 / 27.0 & 89.4 / 35.2 \\
    & KNN & 64.9 / 87.3 & 81.7 / 55.7 & 91.3 / 29.1 & 90.9 / 33.9 & 89.6 / 36.2 & 94.7 / 11.0 & 95.0 / 11.9 & 91.5 / 31.7 & 87.4 / 37.1 \\
    & EBO & 58.8 / 92.3 & 66.0 / 94.2 & 79.3 / 83.6 & 81.2 / 83.7 & 76.5 / 88.8 & 52.7 / 99.0 & 79.6 / 90.0 & 72.5 / 92.3 & 70.8 / 90.5 \\
    & MLS & 64.2 / 91.5 & 72.4 / 93.0 & 85.3 / 73.0 & 83.7 / 78.9 & 81.6 / 85.8 & 58.4 / 99.0 & 84.2 / 86.4 & 72.8 / 93.3 & 75.3 / 87.6 \\
    & SHE & 67.5 / 86.6 & 83.9 / 56.7 & 93.5 / 22.4 & 92.5 / 26.3 & 90.9 / 34.2 & 94.3 / 11.8 & 94.7 / 16.1 & 93.9 / 16.9 & 88.9 / 33.9 \\
    \rowcolor{gray!25}
    & \textbf{SAID} & \textbf{72.6} / \textbf{83.8} & \textbf{87.3} / 47.8 & 96.4 / 15.2 & \textbf{95.3} / \textbf{20.2} & \textbf{92.9} / \textbf{26.9} & \textbf{98.6} / \textbf{3.5} & \textbf{98.2} / \textbf{6.2} & \textbf{98.1} / \textbf{7.7} & \textbf{92.4} / \textbf{26.4} \\

    \hline

    \multirow{8}{*}{\rotatebox[origin=c]{90}{ResNet-50}}
    & VIM  & 65.2 / 80.9 & 78.7 / 61.4 & 88.4 / 31.5 & \textbf{97.6} / 13.0 & \textbf{89.7} / \textbf{33.8} & 99.2 / 2.5 & 97.7 / 6.7 & 98.3 / 4.5 & 89.3 / \textbf{29.3} \\
    & RMDS &   71.1 / 78.4   &   \textbf{81.9} / \textbf{52.7}  &    87.2 / 34.1  &   86.1 / 48.9  &   85.4 / 41.2   &   92.0 /16.6   &   89.1 / 28.8   &   91.9 / 22.0   &   85.6 / 40.3   \\
    & MDS  &   46.9 / 92.9 & 61.3 / 79.8 & 63.3 / 74.4 & 89.3 / 44.8 & 68.0 / 73.6 & 89.6 / 26.4 & 90.6 / 30.9 & 82.9 / 39.2 & 74.0 / 57.8    \\
    & KNN  & 56.4 / 88.3 & 75.8 / 65.2 & 86.5 / 43.6 & 97.4 / 14.7 & 85.3 / 49.6 & \textbf{99.9} / \textbf{0.3} & 97.9 / 6.5 & \textbf{98.5} / \textbf{4.4} & 87.2 / 34.1 \\
    & EBO  & 72.1 / 76.5 & 79.7 / 60.6 & 90.6 / 31.3 & 88.7 / 45.8 & 89.1 / 38.1 & 96.8 / 11.9 & 85.4 / 38.4 & 94.8 / 17.0 & 87.1 / 40.0 \\
    & MLS  & \textbf{72.5} / \textbf{76.2} & 80.4 / 59.5 & 91.2 / 30.6 & 88.4 / 46.1 & 89.2 / 37.9 & 96.1 / 14.0 & 86.5 / 37.9 & 93.1 / 20.2 & 87.2 / 40.3 \\
    & SHE  & 70.4 / 77.4 & 75.9 / 70.8 & \textbf{92.6} / 33.9 & 93.4 / 36.4 & 86.2 / 56.0 & 99.8 / 0.8 & 88.1 / 43.7 & 93.7 / 22.9 & 87.5 / 42.7 \\
    \rowcolor{gray!25}
    & \textbf{SAID} & 64.3 / 84.4 & 78.4 / 63.3 & 92.0 / \textbf{29.1} & \textbf{97.6} / \textbf{11.5} & 87.5 / 45.2 & 99.8 / 0.4 & \textbf{99.6} / \textbf{1.0} & 97.6 / 8.7 & \textbf{89.6} / 30.5 \\
    \hline
\end{tabular}
\end{table*}

\begin{table*}[h]
\caption{\textit{(Left)} \textbf{Covariate shift} detection results for SAID and baseline methods across two models: ViT and ResNet-50. We use ImageNet-1k as the ID dataset and report AUROC $\uparrow$ / FPR95 $\downarrow$ values for three OOD datasets exhibiting covariate shift. Bold indicates the best method. \textit{(Right)} The performance gain in terms of relative AUROC for SAID when using intermediate layers compared to using only the final layer. Using intermediate layers increases AUROC in every dataset.}
\label{tab:covariate_shift_results}
\centering
\scriptsize
\setlength{\tabcolsep}{3pt}
\renewcommand{\arraystretch}{1.05}
\begin{tabular}{|c|c|ccc|c|}
    \hline
     & Method & ImageNet-C & ImageNet-ES & ImageNet-R & Average \\
    \hline

    \multirow{8}{*}{\rotatebox[origin=c]{90}{ViT}}
    & VIM  & 77.5 / 82.3 & 79.4 / 77.3 & 86.1 / 59.7 & 81.0 / 73.1 \\
    & RMDS & 75.5 / 76.5 & 78.2 / 78.9 & 84.5 / 62.3 & 79.4 / 72.6 \\
    & MDS  & 74.7 / 78.1 & 76.5 / 77.8 & 85.9 / 58.4 & 79.0 / 71.4 \\
    & KNN  & 74.3 / 80.9 & 79.4 / 79.0 & 84.3 / 65.5 & 79.3 / 75.2 \\
    & EBO  & 73.4 / 89.7 & 72.7 / 91.0 & 72.5 / 92.3 & 72.9 / 91.0 \\
    & MLS  & 74.4 / 88.8 & 74.6 / 90.3 & 74.8 / 92.1 & 74.6 / 90.4 \\
    & SHE  & 76.6 / 78.0 & 80.3 / 76.4 & 81.3 / 80.0 & 79.4 / 78.1 \\
    \rowcolor{gray!25}
    & \textbf{SAID} & \textbf{80.9} / \textbf{69.8} & \textbf{83.0} / \textbf{71.2} & \textbf{86.6} / \textbf{56.3} & \textbf{83.5} / \textbf{65.8} \\

    \hline

    \multirow{8}{*}{\rotatebox[origin=c]{90}{ResNet-50}}
    & VIM  &   85.1 / 62.5 & \textbf{88.4} / \textbf{45.4} & 81.7 / 66.3 & 85.1 / 58.1   \\
    & RMDS &  79.4 / 71.2 & 81.7 / 59.2 & 72.7 / 73.3 & 77.9 / 67.9 \\
    & MDS  &    68.0 / 76.5 & 69.2 / 71.8 & 49.4 / 88.3 & 62.2 / 78.9  \\
    & KNN  &    82.5 / 70.4 & 85.5 / 57.5 & 82.9 / 69.3 & 83.6 / 65.7   \\
    & EBO  &    82.5 / 72.9 & 87.0 / 53.4 & 84.4 / 71.0 & 84.7 / 65.8 \\
    & MLS  &    82.4 / 72.6 & 86.7 / 53.6 & 84.2 / 71.1 & 84.4 / 65.8   \\
    & SHE  &  80.3 / 74.2 & 75.8 / 73.2 & 81.8 / 77.9 & 79.3 / 75.1 \\
    \rowcolor{gray!25}
    & \textbf{SAID} &  \textbf{87.6} / \textbf{51.5} & 84.8 / 51.9 & \textbf{86.9} / \textbf{60.0} & \textbf{86.4} / \textbf{54.5} \\
    \hline
\end{tabular}
\includegraphics[width=0.45\textwidth,valign=c]{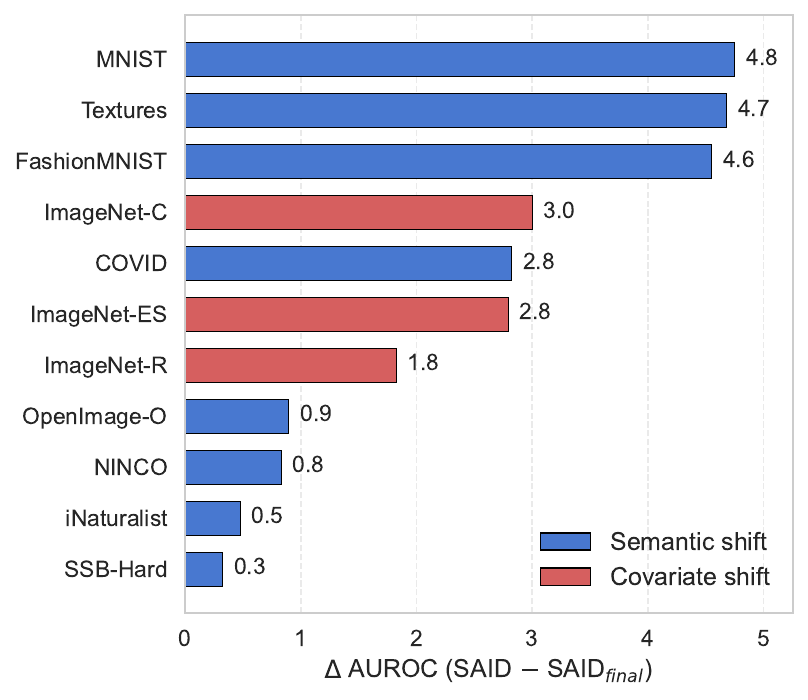}
\end{table*}

\begin{figure*}[t]
    \centering
    \includegraphics[width=0.9\textwidth]{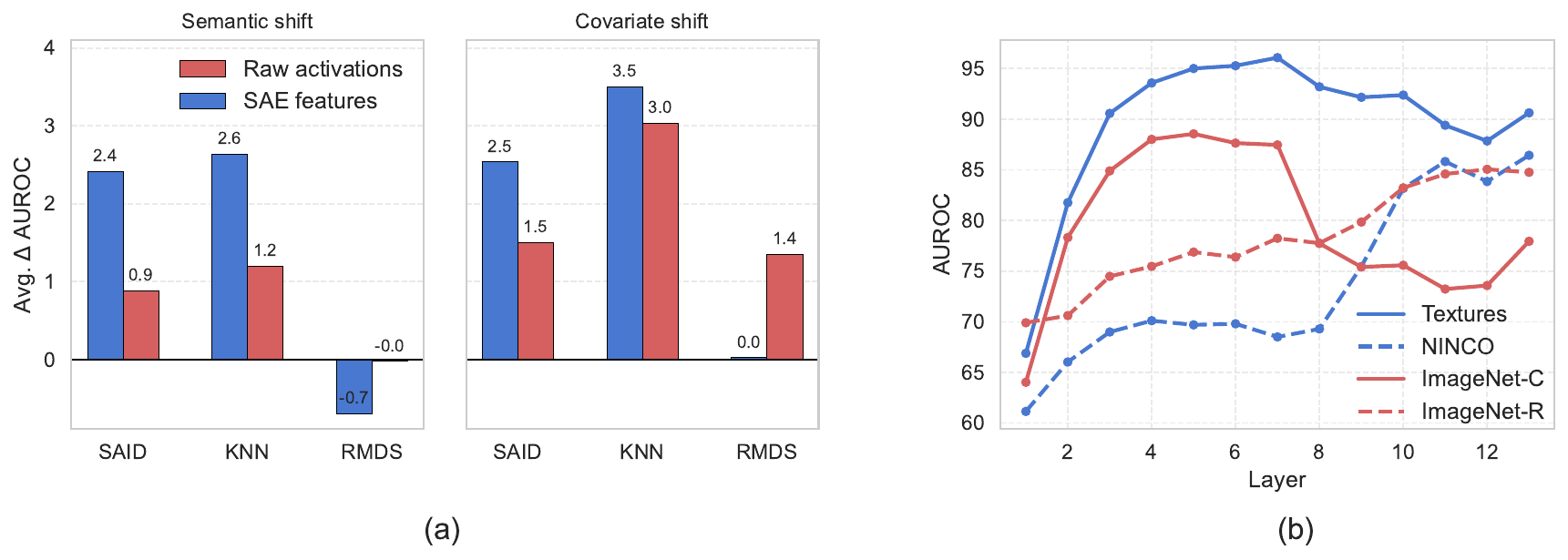}
    \caption{Ablations of intermediate-layer information and layer-wise SAID performance on the ViT backbone. \textit{(a)} For SAID and two competitive baselines, KNN and RMDS, we report the gain in AUROC from adding intermediate layer features on top of the final layer alone. For each method, the left and right bars show the average $\Delta\text{AUROC}$ over the 9 semantic shift and 3 covariate shift OOD datasets, respectively. SAID and KNN both benefit substantially from intermediate layers, while RMDS does not. \textit{(b)} Per-layer AUROC of SAID for 4 OOD datasets, showing that the most discriminative layer varies by dataset. }
    \label{fig:intermediate-ablation-layerwise}
\end{figure*}

\begin{figure*}[t]
\centering
\includegraphics[width=0.9\textwidth]{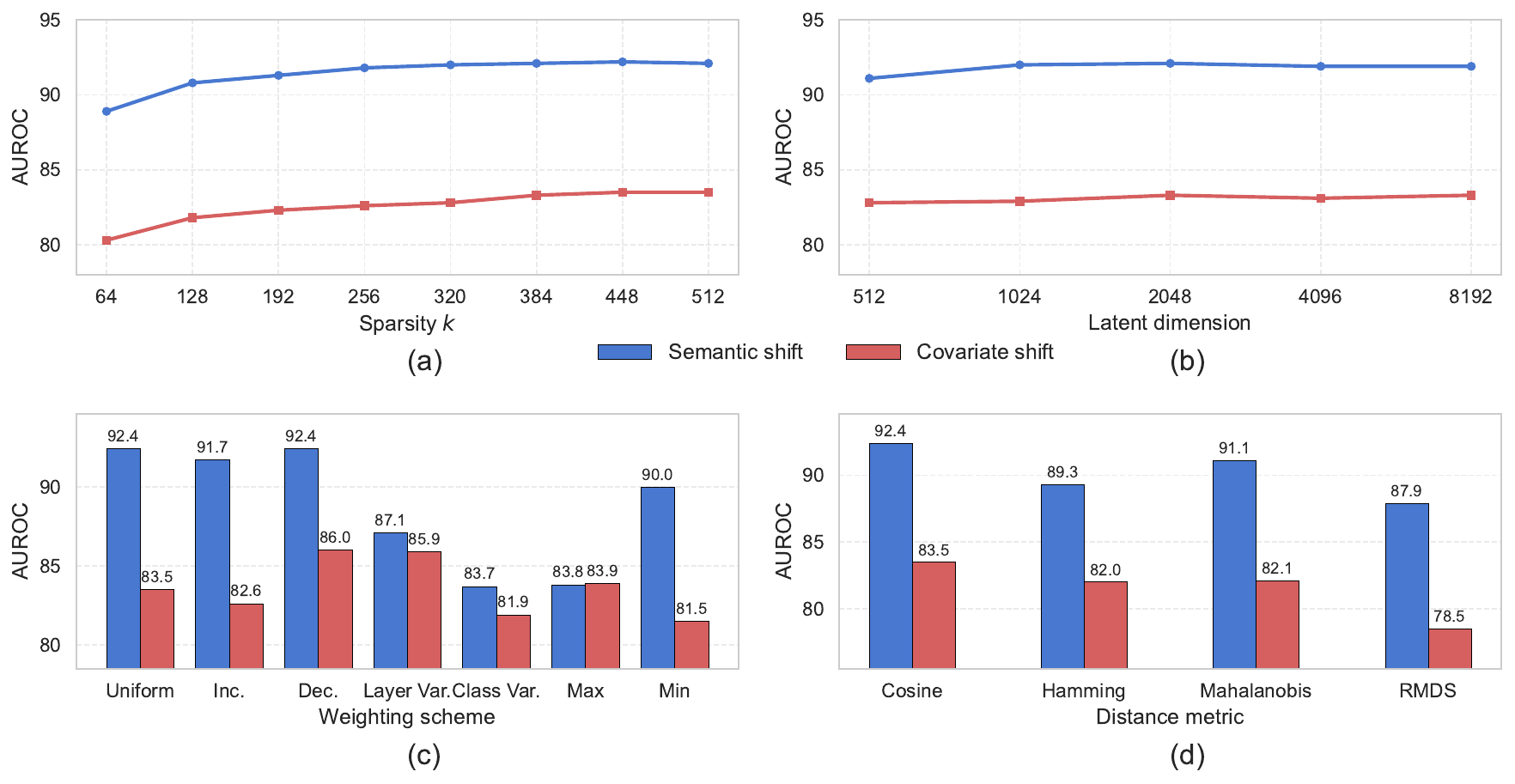} 
\caption{
\textit{(Top Row)} Sensitivity analysis over the sparsity parameter $k$ and the latent dimension of the SAE models.
\textit{(Bottom Row)} Design choice ablations over various layer weighting schemes and distance metrics.
Reported values are AUROC averaged across all datasets in the corresponding benchmark, using ImageNet-1k as the ID dataset with a ViT backbone.
}
\label{fig:sensitivity}
\end{figure*}

\subsection{OOD Detection Performance}

The quantitative results for semantic and covariate shift are shown in Table~\ref{tab:semantic_shift_results} and Table~\ref{tab:covariate_shift_results}, respectively. On both architectures, SAID achieves the best average detection performance on both shift types. 
The improvement is most pronounced for the ViT backbone, where SAID achieves the best average performance on both semantic-shift and covariate-shift benchmarks.
These results suggest that sparse SAE representations provide a strong OOD signal when applied to intermediate representations of transformer-based vision models.
SAID also remains competitive on ResNet-50, achieving the highest average AUROC across both shifts, although the margin over existing methods is smaller.

\subsection{Impact of Intermediate Layers on OOD Detection}

We first ask whether intermediate-layer information is useful at all. Table~\ref{tab:covariate_shift_results}.b shows the AUROC gain from using SAEs trained on multiple intermediate layers rather than only the final layer. Across all datasets, adding intermediate-layer SAEs improves performance, indicating that intermediate representations contain discriminative information that complements the final-layer embedding.

The benefit is largest for datasets that are separable from ID at earlier layers. Figure~\ref{fig:intermediate-ablation-layerwise}.b shows that the most informative layer varies by dataset. Some datasets, such as NINCO and ImageNet-R, perform best at the final layer, while others, such as Textures and ImageNet-C, are better separated in earlier layers. This shows that the discriminative structure for these inputs lives in lower-level features that final-layer methods discard, and motivates aggregating across multiple layers.

A natural follow-up question is whether SAEs are the right way to extract that intermediate-layer information. Figure~\ref{fig:intermediate-ablation-layerwise} shows that both SAID and KNN benefit from adding intermediate layer representations as both SAEs and raw activations, while RMDS performs \textit{worse} in some cases. Although KNN obtains a larger relative gain from both types of representations, SAID achieves stronger absolute performance on both semantic- and covariate-shift benchmarks (Tables~\ref{tab:semantic_shift_results} and~\ref{tab:covariate_shift_results}). SAID is also more compact at inference time: it stores one mean latent vector per class per layer, whereas KNN must store and search a large ID reference set. Together, these results show that intermediate layers provide useful OOD evidence, and that SAE features offer an effective and compact way to aggregate this information.

\subsection{Sensitivity and Ablation Analysis}

We next evaluate the sensitivity of SAID to its main hyperparameters and design choices
(Fig.~\ref{fig:sensitivity}).
Overall, SAID is relatively stable across a range of SAE latent dimensions.
Performance improves when moving from very small latent spaces to moderate-sized
ones, but changes only marginally once the latent dimension is sufficiently large (beyond 1024). 
This suggests that, beyond a certain capacity, additional latent dimensions do not substantially change the concept space used for OOD detection, likely because the SAE already has enough active features to represent the relevant ID structure.
The sparsity level \(k\) has a more pronounced effect.
Increasing \(k\) initially improves performance, indicating that too few active features may underrepresent the visual concepts needed to distinguish among ImageNet classes, though this too saturates beyond 448.

We also ablate how layer-wise scores are aggregated and how similarity is measured in the SAE
latent space.
For score aggregation (Fig.~\ref{fig:sensitivity}.c), we compare uniform averaging against several alternatives: linearly increasing or decreasing weights over layers, variance-based weights computed either globally over ID concept activations or with respect to the predicted class, and max/min aggregation using the largest or smallest layer-wise similarity score. 
Uniform averaging performs competitively across both semantic and covariate shifts, suggesting that a simple layer-agnostic aggregation is a robust default. 
Although some weighting schemes can improve performance for particular shift types, they implicitly assume that certain layers are more informative a priori, requiring either domain expertise or validation samples from the test distribution.

Among distance metrics (Fig.~\ref{fig:sensitivity}.d), cosine similarity performs best overall.
We conjecture that this is the case due to the sparsity of the latent concept vectors.
In this setting, the relative pattern of activated concepts may be more important than the overall activation magnitude.
Cosine similarity captures this by assigning high similarity to inputs with aligned sparse activation patterns, even when their latent vectors differ in scale.

\begin{figure*}[h]
\centering
 \includegraphics[width=0.99\textwidth]{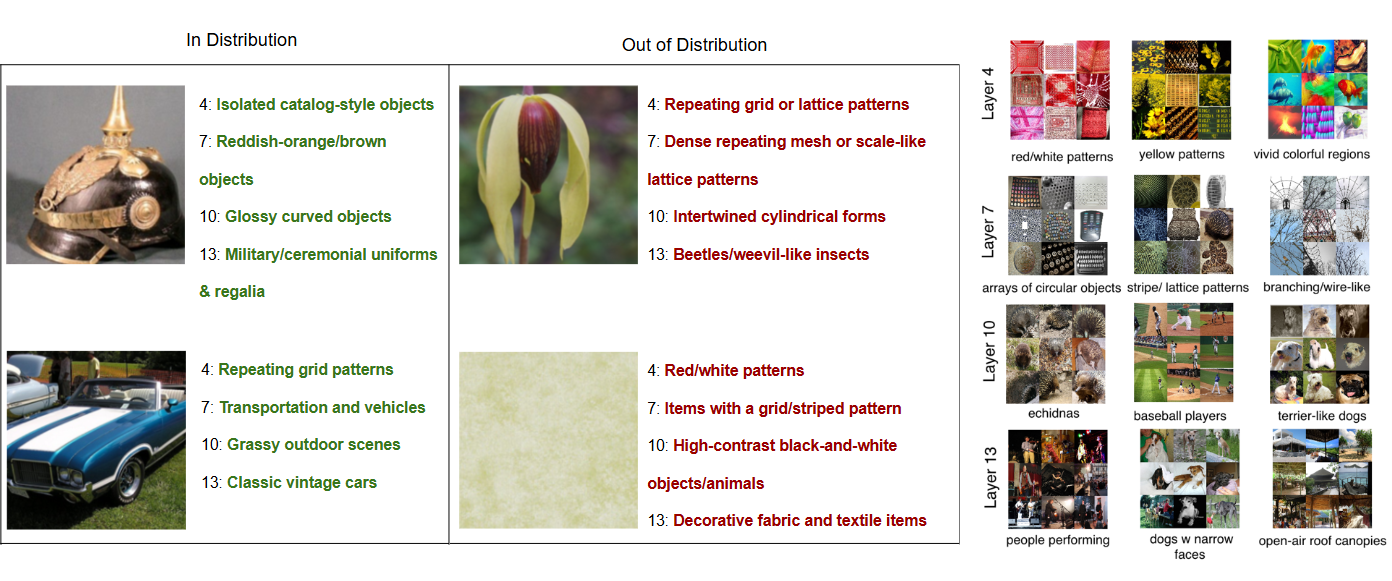}
\caption{
\textit{(Left)} Two ID and two OOD examples, with one of the top‑3 SAE‑derived concept labels for each layer based on activation strength. Concept labels that agree with the ground truth class are colored green; concepts that do not are colored red. ID samples predominantly align with correct concepts, whereas OOD samples trigger mismatched concepts.
\textit{(Right)} For each layer, examples of frequently activated concepts over the ID dataset, labeled by a VLM. Each label is generated by retrieving the 9 images with the highest activation for that concept, presenting them to the VLM, and using its description as the concept label. Later layers tend to yield more abstract, high‑level concepts than earlier ones.
}
\label{fig:concepts}
\end{figure*}

\begin{figure}[t]
\centering
\includegraphics[width=1\columnwidth]{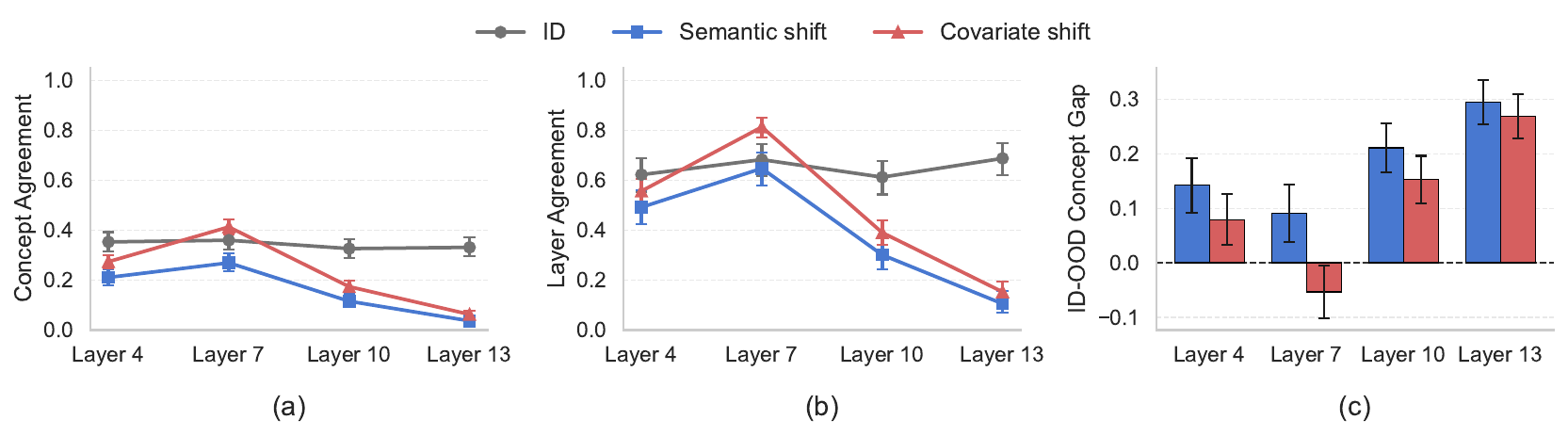} 
\caption{
LLM-judged concept agreement for the ViT model using 200 ID samples, 200 semantic shift OOD samples, and 200 covariate shift OOD samples. 
(a) Shows the fraction of individual SAE concepts judged to be semantically consistent with the input image, i.e. they \textit{agree}.
(b) Shows the fraction of layers that agree with the input image. A layer agrees if at least one of the top-3 concepts agrees. 
(c) The ID--OOD agreement gap highlights layers where distribution shift most strongly reduces concept agreement. 
ID samples exhibit higher concept agreement than OOD samples, indicating that OOD inputs activate concepts that are less semantically consistent with the image.
}
\label{fig:concept_agreement}
\end{figure}

\subsection{Analyzing Concept Activations Under Distribution Shift}

We next evaluate whether SAE-derived concept labels reflect meaningful differences between ID and
OOD inputs.
We construct a test dataset of 400 images: 200 images uniformly sampled from among a subset of 40 classes in the ID dataset, and 200 images uniformly sampled across each OOD dataset in the corresponding shift setting (semantic or covariate).
For each test image, we consider the top activated SAE concepts at each layer and use a
VLM to judge whether each concept is semantically consistent with the image content.
We report two metrics in Fig.~\ref{fig:concept_agreement}: concept agreement, the fraction of activated concepts that agree with the image, and layer agreement, the fraction of layers for which at least one activated concept agrees with the image.

Qualitative examples are shown in Fig.~\ref{fig:concepts}.
ID samples tend to activate concepts that align with the image class or with visually relevant attributes, while OOD samples more often activate concepts that are weakly related or semantically inconsistent.
The quantitative results in Fig.~\ref{fig:concept_agreement} support this trend.
Across layers, ID samples exhibit higher concept agreement than both semantic-shift and covariate-shift OOD samples.
The ID--OOD gap is larger in later layers, suggesting that distribution shift is especially visible in higher-level concept representations.
However, this layer-wise pattern should be interpreted diagnostically rather than causally: disagreement at a later layer may reflect either a mismatch at that layer or the downstream effect of differences that arise earlier in the network.

Overall, these results show that SAE concept labels provide a complementary view of OOD behavior.
While SAID's detection score measures similarity in sparse latent space, concept agreement helps characterize whether the activated latent features correspond to meaningful visual content in the
input.
This makes the learned concepts useful for analyzing how ID and OOD samples differ across
the network's representation hierarchy.

\section{Discussion and Conclusion}

We introduced \textbf{S}parse \textbf{A}utoencoders for \textbf{I}nterpretable \textbf{D}etection (SAID), a post-hoc OOD detection framework that applies sparse autoencoders to intermediate representations of frozen vision models.
By comparing sparse latent encodings to class-conditional ID prototypes, SAID provides a simple similarity-based OOD score that leverages information distributed across the network rather than relying only on final-layer features.
Empirically, SAID achieves strong performance across both semantic and covariate distribution shifts on ViT and ResNet-50 backbones.
Our results show that intermediate representations contain discriminative structure that is often reduced or obscured in final-layer embeddings, and that SAE features provide an effective way to extract and aggregate this information for OOD detection.

Beyond detection performance, SAID provides a concept-level view of how ID and OOD inputs are represented across layers.
Using VLM-labeled SAE features, we find that ID samples tend to activate concepts that are more semantically consistent with the input image than those activated by OOD samples.
This suggests that SAE concepts can serve as a useful diagnostic tool for analyzing how distribution shift affects learned representations.
Future work could extend this framework to larger models, language and multimodal domains, and more natural forms of distribution shift.

\textbf{Limitations:}
Despite promising results, our approach has limitations, the most significant of which is that we have limited our current evaluation to vision domains.
Methodologically, the use of a fixed-$k$ sparsity constraint, where a constant number of features are active for any given input, may be a restrictive assumption. The inherent complexity of data varies; some ID inputs might be adequately described by very few concepts, while others may require more. A more flexible sparsity mechanism could potentially improve representation quality and, consequently, OOD detection performance.

Furthermore, the computational cost associated with the proposed method is non-trivial. 
Training a separate, high-capacity SAE for each of the intermediate layers is a resource-intensive process. This requirement for substantial computational power and time could pose a barrier for applications involving extremely large models or environments with limited hardware resources.

\bibliographystyle{plainnat}
\bibliography{references}


\appendix

\section{VLM-Based Concept Annotation Prompt}


\begin{tcolorbox}[
    colback=gray!10,
    colframe=black,
    arc=4pt,
    boxrule=0.5pt
]
\footnotesize
\textbf{[SYSTEM]}
We are studying features of a sparse autoencoder (SAE) trained on ViT-B/16 activations. Each SAE feature (neuron) looks for some particular thing in an image. Look at the images that the feature activates for and summarize in a single sentence what the feature is looking for.
You will be shown a grid of the top-activating images followed by activation strengths (same grid order, left to right, top to bottom). Activation values range from 0 to 10. A non-zero activation value means the feature found what it is looking for. The higher the activation value, the more strongly that image activated the feature.

\vspace{8pt}
\textbf{[USER]}

\smallskip
Neuron 300 Activations:

\smallskip
Images arranged as a 3$\times$3 grid (left$\to$right, top$\to$bottom). Each cell shows the original image.

\smallskip
\textit{[IMAGE: 3$\times$3 grid]}

\smallskip
Activation strengths (same grid order, left$\to$right, top$\to$bottom):
\begin{quote}
$[10,\ 10,\ 9]$ \\
$[9,\ 9,\ 9]$ \\
$[9,\ 9,\ 9]$
\end{quote}

Explanation of neuron 300 behavior: the main thing this neuron does is find

\smallskip
\textbf{[ASSISTANT]} white and brown dogs.

\vspace{8pt}
\textbf{[USER]}

\smallskip
Neuron 100 Activations:

\smallskip
Images arranged as a 3$\times$3 grid (left$\to$right, top$\to$bottom). Each cell shows the original image.

\smallskip
\textit{[IMAGE: 3$\times$3 grid]}

\smallskip
Activation strengths (same grid order, left$\to$right, top$\to$bottom):
\begin{quote}
$[10,\ 9,\ 9]$ \\
$[9,\ 9,\ 9]$ \\
$[9,\ 9,\ 9]$
\end{quote}

Explanation of neuron 100 behavior: the main thing this neuron does is find

\smallskip
\textbf{[ASSISTANT]} vibrant yellow and orange colors.

\vspace{8pt}
\textbf{[USER]}

\smallskip
Neuron 223 Activations:

\smallskip
Images arranged as a 3$\times$3 grid (left$\to$right, top$\to$bottom). Each cell shows the original image.

\smallskip
\textit{[IMAGE: 3$\times$3 grid]}

\smallskip
Activation strengths (same grid order, left$\to$right, top$\to$bottom):
\begin{quote}
$[10,\ 9,\ 9]$ \\
$[8,\ 8,\ 8]$ \\
$[8,\ 7,\ 7]$
\end{quote}

Explanation of neuron 223 behavior: the main thing this neuron does is find

\smallskip
\textbf{[ASSISTANT]} spiders on a spider web.

\vspace{8pt}
\textbf{[USER]}

\smallskip
Neuron 65 Activations:

\smallskip
Images arranged as a 3$\times$3 grid (left$\to$right, top$\to$bottom). Each cell shows the original image.

\smallskip
\textit{[IMAGE: 3$\times$3 grid]}

\smallskip
Activation strengths (same grid order, left$\to$right, top$\to$bottom):
\begin{quote}
$[10,\ 10,\ 10]$ \\
$[10,\ 9,\ 9]$ \\
$[9,\ 9,\ 9]$
\end{quote}

Explanation of neuron 65 behavior: the main thing this neuron does is find

\end{tcolorbox}



\section{VLM Concept-Image Agreement Prompt}
\begin{quote}\begin{scriptsize}\begin{verbatim}
def get_gpt_verdict(image_b64: str, concepts: list[dict]) -> Union[dict, str]:
    """
    Sends the image and all concepts to GPT-5.5 in a single call.
    Asks GPT to return a JSON object mapping each concept string to true/false.

    Returns a dict like:
        {"giant pandas": true, "grid or mesh-like patterns": false, ...}
    or a string error tag on failure.
    """
    if not concepts:
        return "no_concepts_found"

    concept_strings = [c["concept"] for c in concepts]
    numbered_list = "\n".join(f"{i+1}. {c}" for i, c in enumerate(concept_strings))

    prompt_text = (
        "I will provide you an image and a set of visual concepts in natural language. "
        "Consider each concept and decide whether it could be applied to the image. "
        "These concept labels may be abstract in nature. "
        "Err on the side of caution — only mark a concept false if it clearly does not describe "
        "the image or references visual features that are not present at all.\n\n"
        f"Concepts:\n{numbered_list}\n\n"
        "Respond with a single JSON object where every key is the exact concept string "
        "(copied verbatim from the list above) and the value is true or false. "
        "Return only valid JSON — no markdown, no explanation, no extra text."
    )

    messages = [
        {
            "role": "system",
            "content": (
                "You are a judge for a machine learning experiment evaluating visual concept labels. "
                "You respond only with a valid JSON object mapping concept strings to boolean values."
            ),
        },
        {
            "role": "user",
            "content": [
                {"type": "text", "text": prompt_text},
                {"type": "image_url", "image_url": {"url": f"data:image/png;base64,{image_b64}"}},
            ],
        },
    ]
    ...

\end{verbatim}
\end{scriptsize}\end{quote}

\section{SAE training hyperparameters}

\begin{table}[h!]
\caption{Hyperparameters used to train SAEs for ViT}
\centering
\small\selectfont
\begin{tabular}{@{}lc@{}}
\hline
\textbf{Hyperparameter} & \textbf{Value} \\
\hline
Latent dimensions & 2048 \\
k & 128, 384 \\
Epochs & 30 \\
Learning rate & 0.0001 \\
Optimizer & Adam \\
Training batch size & 4096 \\
Weight initialization scheme & Kaiming \\
Tied weights & True \\
\hline
\end{tabular}
\label{tab:sae_hyperparameters}
\end{table}


\begin{table}[h!]
\caption{Hyperparameters used to train SAEs for ResNet-50}
\centering
\small\selectfont
\begin{tabular}{@{}lc@{}}
\hline
\textbf{Hyperparameter} & \textbf{Value} \\
\hline
Latent dimensions & 1024 \\
k & 64, 96 \\
Epochs & 30 \\
Learning rate & 0.0001 \\
Optimizer & Adam \\
Training batch size & 4096 \\
Weight initialization scheme & Kaiming \\
Tied weights & True \\
\hline
\end{tabular}
\label{tab:sae_hyperparameters_2}
\end{table}

Table~\ref{tab:sae_hyperparameters} and Table~\ref{tab:sae_hyperparameters_2} lists the hyperparameters we used to train all SAEs for our method with ViT and ResNet-50 as the backbone architectures respectively. The k hyperparameter is the k value used in the top-k function. The For ViT, the first k value corresponds to the k used at Layers 4 and 7, whereas the second is the k used at Layers 10 and 13. For ResNet-50, the first k value corresponds to the k used at Layers 10 and 22, whereas the second is the kused at Layers 40 and 49. Tied weights refers to initializing each SAE decoder weight matrix to the transpose of the encoder weight matrix (but not coupled during training). The SAEs were trained on A100 GPUS using 40GB of memory. Software libraries we used include Pytorch 2.7.0 and OpenOOD 1.5.



\end{document}